\documentclass[conference]{IEEEtran}
\IEEEoverridecommandlockouts
\usepackage{cite}
\usepackage{amsmath,amssymb,amsfonts}
\usepackage{algorithmic}
\usepackage{graphicx}
\usepackage{textcomp}
\usepackage[x11]{xcolor}

\usepackage{hyperref}
\usepackage{multirow}
\usepackage{algorithm}
\usepackage{booktabs}
\usepackage{algorithmic}
\usepackage{here}
\usepackage{colortbl}
\usepackage{flushend}

\colorlet{RemainHigh}{green!10}
\colorlet{Degraded}{red!10}
\colorlet{RemainLow}{yellow!30}
\colorlet{Improved}{cyan!10}

\def\BibTeX{{\rm B\kern-.05em{\sc i\kern-.025em b}\kern-.08em
    T\kern-.1667em\lower.7ex\hbox{E}\kern-.125emX}}

\def\topN{\textit{topN}}
\def\confThold{\textit{confTh}}

\begin{document}

\title{Transductive Data Augmentation with Relational Path Rule Mining for Knowledge Graph Embedding
}

\author{\IEEEauthorblockN{Yushi Hirose$^1$, Masashi Shimbo$^{2,3}$, Taro Watanabe$^4$}
\IEEEauthorblockA{\textit{$^1$Tokyo Institute of Technology, Tokyo, Japan} \\
\textit{$^2$Chiba Institute of Technology, Chiba, Japan}\\
\textit{$^3$Riken AIP, Tokyo, Japan}\\
\textit{$^4$Nara Institute of Science \& Technology, Nara, Japan}\\
hirose.y.am@m.titech.ac.jp, shimbo@stair.center, taro@is.naist.jp}
}

\maketitle

\begin{abstract}
For knowledge graph completion, two major types of prediction models exist: one based on graph embeddings, and the other based on relation path rule induction.
They have different advantages and disadvantages.
To take advantage of both types, hybrid models have been proposed recently. One of the hybrid models, UniKER, alternately augments training data by relation path rules and trains an embedding model.
Despite its high prediction accuracy, it does not take full advantage of relation path rules, as it disregards low-confidence rules in order to maintain the quality of augmented data.
To mitigate this limitation, we propose transductive data augmentation by relation path rules and confidence-based weighting of augmented data.
The results and analysis show that our proposed method effectively improves the performance of the embedding model by augmenting data that include true answers  or entities similar to them.
\end{abstract}


\section{Introduction}
A \emph{knowledge graph} (KG) is a labeled directed graph to store facts about the world.
Its nodes represent entities, and edges represent relations between them,
with the label of each edge indicating the type of relation.
For example, the fact that Bob was born in New York City is represented by an edge labeled \textit{bornIn} from node \textit{Bob} to node \textit{NYC}.
An edge, as well as the proposition it signifies, can be denoted by a \emph{triplet}, such as $(\textit{Bob}, \textit{bornIn}, \textit{NYC})$;
the first and the last components are entities, respectively called head and tail, and the second component is the type of relation.
KGs have many applications 
such as Q\&A~\cite{QA} and entity linking~\cite{EL} in natural language processing.

KGs are usually incomplete;
i.e., 
some fact triplets are missing, 
although their component entities and relation types 
are present in other triplets.
The task of detecting such missing facts is called \emph{knowledge graph completion} (KGC).
For KGC, two major approaches exist: embedding models~\cite{rescal,complex,tucker}, and relation path rule models~\cite{rulen}.
Embedding models learn feature vectors (or, in some cases, matrices) of entities and relations from existing facts.
They then predict the truth value of missing triplets by the score computed with mathematical operations on the feature vectors.
By contrast, relation path rule models predict the truth value of missing triplets on the basis of the relation sequences (relation paths) between entities,
regarding them as logical rules.

Both approaches have advantages and disadvantages.
Embedding models can learn features from entities and relations in the entire KG, 
but they do not explicitly model relation path information that can be potentially useful.
Relation path rule models can induce logical rules intrinsic in a KG, which contributes to generating fewer false positives.
However, they cannot predict triplets if there are no relation paths between entities.

To take advantage of the two approaches, various hybrid models~\cite{traversing,kblrn,fb15k237,uniker} have been proposed.
UniKER \cite{uniker} is one of them. 
This method alternates between (i) augmenting the training set with triplets obtained by relation path rules and (ii) training an embedding model with the augmented training data.
It has been shown that UniKER achieves high KGC accuracy.
However, to avoid producing incorrect triplets for augmentation, it uses only a few relation path rules that are found to be highly reliable.
It also needs multiple passes of augmenting data and training embedding models to reduce data noise.

In this paper, we propose a simple method that augments data by a large number of relation path rules and uses embedding models for prediction after a single training pass.
As we see in Section~\ref{sec:exp}, randomly augmenting data does not improve prediction accuracy. 
To augment useful data for test queries, we consider a transductive learning setting~\cite{transductive}, which assumes test queries (but not their answers) are known at the time of training \cite{transductive}.
Transductive learning a form of semi-supervised learning originally studied in the context of support vector machines~\cite{transductive,tsvm}.
In more recent years, it has been actively investigated with zero-/few-shot tasks (e.g., \cite{Kodirov+:2015:ICCV,Guo+:2016:AAAI,Boudiaf+:2020:NeurIPS,Cui+:2021:ICML}), where the system has to adapt to unseen classes that can have different data distributions from classes observed in training.
To the best of our knowledge, this paper is the first to investigate the effect of a transductive setting in KGC.
It is a problem setting easier than the standard KGC, but the proposed method obtains useful data effectively and achieves a prediction accuracy comparable with UniKER.

\section{Related Work}
The task of KGC is formalized as follows:
Let a KG be given, with $E$ the set of entities in the KG and $R$ the set of relation types.
Given a query (test) triplet $(e,r,?)$ or $(?,r,e)$ where $e \in E$ and $r \in R$, a system must rank all entities in $E$ by their plausibility as the answer to the query;
i.e., by how likely the triplet is to be a fact when each entity is slotted in the position of `$?$'.
Thus, the task boils down to defining a score function over triplets that indicates their likelihood.
Mainly, three types of models have been proposed thus far: A. embedding models, B. relation path rule models, and C. hybrid models.

\subsection{Embedding models}
Embedding models define the score of a triplet in terms of mathematical operations on feature vectors (or matrices) of entities and relations.
Triplets with higher scores are considered more likely to be true triplets.
RESCAL~\cite{rescal} is the first bilinear KG embedding model and TuckER~\cite{tucker} is a recently proposed model shown to perform better.
The score functions for the respective models are shown as $f_{\rm{RESCAL}}$ and $f_{\rm{TuckER}}$ in Eqs.~\eqref{eq:RESCAL} and \eqref{eq:TuckER}.
\begin{align}
  \label{eq:RESCAL}
  f_{\rm{RESCAL}}(e_h,r,e_t) &=\mathbf{e}^{\top}_{h} \mathbf{R} \mathbf{e}_{t}\\
  \label{eq:TuckER}
  f_{\rm{TuckER}}(e_h,r,e_t) &=\mathbf{W} \times_{1} \mathbf{e}_{h} \times_{2} \mathbf{r} \times_{3} \mathbf{e}_{t}
\end{align}
where
$\times_{n}$ denotes the tensor $n$-mode product operation,
$\mathbf{e}_h$ and $\mathbf{e}_t$ denote the embedding vectors of head and tail entities $e_h$ and $e_t$, $\mathbf{r}$ and $\mathbf{R}$ are the embedding vector and matrix for relation $r$,
and $\mathbf{W}$ is the core tensor common to all triplets.
These vectors and matrices are usually trained by stochastic gradient descent or one of its variants.

\subsection{Relation path rule models}

A \emph{relation path} is a sequence of relations.
That is, if $ r_1, \ldots, r_n$ are relations, then $p \equiv (r_1, \ldots, r_n)$ is a relation path of length $n$.
We say relation path $p = (r_1, \ldots, r_n)$ exists in a KG from node $e_0$ to node $e_n$ if for some entities $e_1, \ldots, e_{n-1}$, there exist triplets $(e_0,r_1,e_1), (e_1,r_2,e_2), \ldots ,(e_{n-1},r_n,e_n)$ in the KG.
Further, we define $p(e_0, e_{n})$ to be true if relation path $p$ exists from $e_0$ to $e_n$; otherwise it is false.

To see the utility of relation paths in KGC, consider the query (\textit{Bob}, \textit{nationality}, ?).
If (\textit{Bob}, \textit{bornIn}, \textit{NYC}) and (\textit{NYC}, \textit{isCityOf}, \textit{USA}) are in a KG,
we can infer that \textit{USA} should be the answer for the query.
That is, the existence of the relation path (\textit{bornIn}, \textit{isCityOf}) helps answer queries about relation \textit{nationality}.

Generalizing the inference pattern illustrated above,
we can state the relation path rule to predict relation $r$ from path $p = (r_1,r_2, \ldots ,r_n)$
by the first order predicate logic as follows:
\begin{multline}
\label{pathrule}
  \exists e_0,e_1, \ldots, e_{n}:
  (e_0,r_1,e_1)\land 
  \ldots \land(e_{n-1},r_n,e_n) \\
  \rightarrow (e_0,r,e_n)
\end{multline}
The confidence of this rule can be estimated from the frequency of triplets and relation paths in the KG,
specifically by Eq.~\eqref{eq:pcwa}, based on the partial closed world assumption~\cite{amie}.

\begin{equation}
  \label{eq:pcwa}
  \operatorname{Conf}_r (p)= 
  \frac{
    \left|
      \{ (e_0, e_n) \mid p(e_0, e_n) \land (e_0, r, e_n) \}
    \right|
  }{
    \left|
      \{ (e_0, e_n) \mid \exists e': p(e_0, e_n)\land (e_0, r, e') \}
    \right|
  }
\end{equation}
where triplets should be interpreted as true if they are present in the KG, or false otherwise.
Given a query, entities predicted by higher confidence rules are ranked higher as the answer.

\subsection{Hybrid models}

Embedding models can learn features of entities and relations from the entire KG, without explicitly considering relation path information.
Relation path rule models can use logical rules and have fewer false positives, but they cannot predict the triplets without relation paths between entities.
Hybrid approaches that combine the two types of models have been proposed recently.
One of the hybrid models\cite{traversing} use relation path features to regularize KGE learning.
Meilicke et al.~\cite{rulen} proposed ensemble learning of embedding and relation path rule models. 

UniKER~\cite{uniker} is an approach that uses relation path rules for data augmentation.
It first augments the training set using relation path rules mined from the given KG, and then trains an embedding model using the augmented data.
To maintain the quality of augmented triplets, UniKER uses only a few rules that it deems highly reliable.
After training an embedding model, triplets augmented by the rules but has a low embedding score by the embedding models are regarded as noise, and filtered out.
The new training set is used for the next data augmentation process and these steps are repeated.

The augmented data improves basic embedding models, and the prediction accuracy of UniKER is comparable to the state-of-the-art methods in KGC.
UniKER takes a cautious approach in data augmentation to avoid incorrect triplets from being injected to the training set. 
It thus uses only a few relation path rules with the highest confidence scores in an iteration, and repeats data augmentation and noise reduction process. 
These limit the usage of low-confidence rules and also hampers computational efficiency.

\section{Transductive Data Augmentation and Confidence Weighting}

\subsection{A transductive setting for data augmentation}
  
We propose transductive data augmentation for KGC to mitigate the disadvantages of UniKER.
Previous studies in KGC did not explore transductive settings, and this is the first study to investigate this direction.

Transductive learning \cite{transductive} is a problem setting such that test queries are provided at the time of training along with the training data.
Unlike the standard setting of inductive learning,
learning a model that can answer general queries is not requested; it suffices to merely provide answers to those specific queries given at the training time.

Thus, in the transductive setting for KGC, test queries of form $(e_h,r,?)$ or $(?,r,e_t)$
are available
at the time of training.
This deviates from the standard KGC setting, but it is not an unrealistic scenario.
For instance, suppose that a human editor of a KG knows in advance that triplets for particular relations are missing more than other relations,
in which case she might want to focus on the completion of these specific relations.
Indeed, in existing KBs (e.g., FB15k), there is a huge difference in the number of triplets across relations.

Unlike UniKER, which focus on not introducing incorrect data into the augmented training set, we harvest a wider variety of potentially informative data about true answers. 
Given a test query $(e_h,r,?)$ (or $(?,r,e_t)$),
we use the relation path rules to predict the potential answer entity $e_t'$ (or $e_h'$).
We then add the triplet $(e_h,r,e_t')$ (or $(e_h',r,e_t))$
to the training data.
We expect to augment the data of true answer or the data where the potential answer entity is similar to the true answer. These augmented data are assumed to be useful to answer test queries.
We train an embedding model with the augmented data once and directly use the embedding model to predict test query answers without other processes used in UniKER.

In UniKER, augmented triplets are considered as definite truth, in the same way as the triplets originally in the KG.
We instead weight augmented triplets by the confidence values of the rules used to predict the triplet.
This allows us to consider uncertainty of heuristically augmented data, and thus to take into account relation path rules that have low confidence scores.

\subsection{Proposed data augmentation framework}
\label{sec:architecture}

Given a KG (consisting of triples in the training set) and a set of queries $Q$, our method takes the following steps.
\begin{enumerate}
\item Mine relation path rules from the KG
\item Apply mined rules to generate candidate triplets for augmentation
\item Filter candidate triplets
\item Train an embedding model with the triplets in KG and the triplets filtered in the previous step.
\end{enumerate}
We describe the first three steps below.
The last step is omitted, because it is the standard procedure for training a KG embedding model, except for the additional use of augmented triplets.

\paragraph*{$1)$ Mine relation path rules}
\label{sec:mining-rules}

We use the subgraph feature extraction method \cite{sfe} to extract relation path rules from the KG.
To be precise, for each relation $r$ in the KG, we run the following steps.
First we sample a set of triplets with relation $r$ from the KG, and repeat the following process for each sampled triplet $(e_h,r,e_t)$.
Bidirectional random walks are performed in the KG starting from nodes $e_h$ and $e_t$ up to a predetermined length $L$.
Whenever the end nodes of the bidirectional walks meet, two traversed paths are concatenated to produce a full relation path from $e_h$ to $e_t$,
which is then harvested as a rule to predict relation $r$.
After all rules are harvested, we compute their confidence scores by Eq.~\eqref{eq:pcwa}.
The details of the mining settings can be found in Appendix~\ref{sec:rule-mining-details}.
To reduce computational costs in the subsequent steps, for each relation, only rules with the highest 1000 confidence scores are retained. 
It should be noted that this number of rules is much larger than that used by UniKER, which is only several rules per relation.

\paragraph*{$2)$ Apply mined rules to generate candidate triplets}
\label{sec:applying-rules}

For each query $(e_h,r,?)$ in the query set $Q$, 
the following is applied.
For every entity $e \in E$ such that $(e_h, r, e)$ does not exist in the KG,
we check whether any harvested rule for $r$ is applicable to triplet $(e_h, r, e)$; i.e., if a relation path in a rule for $r$ exists from node $e_h$ 
to node $e$ in the KG.
If an applicable rule is found, $(e_h,r,e)$ 
is added to the candidate set of triplets to be considered for augmentation. 
The added triplet is associated with a \emph{confidence weight} equal to the confidence score of the rule that produced it.
If two or more rules are applicable for $(e_h, r, e)$, its weight is set to the highest confidence score among the applicable rules.
Queries of form $(?, r, e_t)$ can be processed in a similar way.

We note that the existence of relation paths can be checked efficiently by sparse graph adjacent matrix multiplication \cite{uniker}.

\paragraph*{$3)$ Filter candidate triplets for augmentation}
\label{sec:tholdaug}

The candidate triplets generated in Step~$2$ 
are filtered with the following two criteria: 
\begin{itemize}
\item 
  First, triplets whose confidence weights are less than $\confThold$ are removed.
\item 
  Next, for each query $(e_h, r, ?)$ in $Q$, we filter out triplets matching the query with lower confidence scores, so that at most $\topN$ triplets of form $(e_h, r, e)$, $e\in E$ are in the final augmented training set.
  Note that this final set contains not only the triples augmented from the candidate set but also the original training triplets.
  Thus, if $\topN$ or more triplets of form $(e_h, r, ?)$ are already present in the original training set, no triplets for the query $(e_h, r, ?)$ are augmented.
\end{itemize}
Triplets surviving these filters are used for augmenting the training data.


\section{Experiments}
\label{sec:exp}

We evaluate the KGC performance of the proposed method in the experiments.
We also execute an augmentation method without transductive setting because the data augmentation can be done with randomly generated queries (See the details in Section~\ref{sec:results}).

\subsection{Experimental settings}
\label{sec:exp-settings}

\subsubsection{Datasets}

Two standard KGC datasets, WN18RR~\cite{wn18rr} and FB15k-237~\cite{fb15k237}, are used for evaluation.
Their statistics are shown in Table~\ref{tab:dataset}.

\begin{table}[tb]
  \caption{\label{tab:dataset} dataset statistics}
  \centering
  \small
  \begin{tabular}{l r r r r r }\toprule
              &        &       & \multicolumn3c{\# of triplets} \\
    \cmidrule{4-6}
    Dataset   & $|E|$  & $|R|$ & training & validation & test   \\\midrule
    WN18RR    & 40,943 & 11    & 86,835   & 3,034      & 3,134  \\
    FB15k-237 & 14,541 & 237   & 272,115  & 17,535     & 20,466 \\ \bottomrule
  \end{tabular}
  \small
  
\end{table}

\subsubsection{Evaluation metrics}

We use Mean Reciprocal Rank (MRR) and Hits@10 as evaluation metrics.
Hits@10 is the proportion of test queries whose true answer is ranked in the top 10 in the rankings.
We use the ``filtered'' ranking setting~\cite{transe};
when multiple correct answers (entities) exist for a query generated by masking one of the entities in a gold triplet,
all these correct entities are excluded from the ranking except for the one in the original triplet, so that there is always a single correct answer for any given query.

For each test triplet $(e_h,r,e_t)$ in the given dataset, a test query $(e_h,r,?)$ is generated, and we evaluate the rank of true answer $e_t$.

\begin{table*}[!t].
  \caption{\label{kgcresult} Experimental results of KGC with each model. Bold numbers denote the best scores in each column. Underlined numbers denote the best scores in each column of separated sections.}
  \centering
  \small
  \begin{tabular}{l l c c c c}\toprule
                            &                             & \multicolumn{2}{c}{WN18RR} & \multicolumn{2}{c}{FB15k-237}                                      \\ \cmidrule(r){3-4} \cmidrule{5-6}
   Embedding model          & Data augmentation approach  & MRR                        & Hits@10                    & MRR               & Hits@10           \\ \midrule
    \multirow{3}{*}{RESCAL} & Base (no data augmentation) & 0.463                      & 0.518                      & 0.351             & 0.536             \\
                            & Random Augmentation         & 0.456                      & 0.513                      & 0.351 & 0.535             \\
                            & Transductive Augmentation   & \underline{0.497}          & \underline{0.565}          & \underline{0.355} & \underline{0.544} \\ \midrule
    \multirow{3}{*}{TuckER} & Base (no data augmentation) & 0.464                      & 0.517                      & 0.354             & 0.540             \\
                            & Random Augmentation         & 0.453                      & 0.503                      & \underline{0.357}             & 0.541 \\
                            & Transductive Augmentation   & \textbf{\underline{0.508}} & \textbf{\underline{0.573}} & 0.356 & \underline{0.542} \\ \midrule
    DistMult                & UniKER\cite{uniker}     & 0.485                      & 0.538                      & \textbf{0.533}    & \textbf{0.587}    \\
    \bottomrule
  \end{tabular}
\end{table*}

\subsubsection{Compared methods}

We compare embedding models trained with three different data augmentation types: ``Base'', ``Random Augmentation'' and ``Transductive Augmentation.''
\begin{itemize}
\item 
  With ``Base,'' no data augmentation is done to the training set.
  This is the baseline.
\item 
  With ``Transductive Augmentation,'' the training set is augmented by the procedure of Section~\ref{sec:architecture}.
  This is the proposed method.
  For hyperparameter tuning, we first generate the query set $Q$ from the validation set, and train a model by the procedure of Section~\ref{sec:architecture}.
  After hyperparameters are set, we discard the trained model, re-generate queries $Q$ now from the test set, generate a new set of augmented training set, and train the embedding model from scratch.
  
  Notice that the triplets in the test set are used only to generate queries (in the same way as used at the test time), and test triplets
  (i.e., the correct answers to the generated queries) are not used directly for training. 
\item 
  With ``Random Augmentation,'' the training set is augmented by the same method as ``Transductive Augmentation,'' except that the input queries are generated by random sampling as follows:
  We sample one entity uniformly at random from the entity set and one relation proportionally to the number of relation triplets in the training set.
  We repeat sampling queries until the total number of queries reach that used for ``Transductive Augmentation.''
\end{itemize}

In both Transductive and Random Augmentation, rules are mined from the KG consisting of the triples in the training set; the validation and test sets are not used for rule mining.
For the embedding models, we use RESCAL and TuckER.

\subsubsection{Training settings}
\label{sec:training-settings}

We set hyperparameters following the previous work \cite{olddog}, 
either with or without augmented data.
For data augmentation, relation path rules of length 1 to 6 are mined and applied; see Section~\ref{sec:mining-rules},$(1)$ and Appendix~\ref{sec:rule-mining-details}.
The threshold values for data augmentation are set by grid search over $\topN\in\{5,50\}$ and $\confThold\in \{0, 0.6\}$.
These threshold values and early stopping of the training are determined by the MRR on the validation data. 
See Appendix~\ref{sec:hyperparams} for the tuned values of hyperparameters and the selected thresholds.

For each relation $r$, we introduce its ``reciprocal'' relation $r^{-1}$, and for each triplet $(e_h,r,e_t)$ in the KG, we add $(e_t,r^{-1},e_h)$ also in the KG.
Adding these triplets balances the number of occurrences of head and tail entities in the KG, and is reported to improve prediction accuracy~\cite{canonical}.

We use the binary cross-entropy loss and the KvsAll~\cite{olddog} method for generating negative examples.
In KvsAll, one mini-batch instance is a pair of a head entity and a relation, $(e_h,r)$.
The loss function for this pair is: 

\begin{multline}
  L = -\frac{1}{|E|}
  \sum_{e'\in E}
  \left(
    y_{e'}
    \log \sigma \left( f(e_h,r,e') \right) \right .\\
  \left.
    +\left( 1- y_{e'} \right) 
    \log
    \left(
      1-\sigma\left( f(e_h,r,e') \right)
    \right)
  \right) ,
\end{multline}
where $E$ is the set of all entities in the KG, $f$ represents the score function of the model, and $\sigma$ is the sigmoid function.
The label $y_{e'} = 1$ if $(e_h,r,e')$ is originally in the KG.
If $(e_h,r,e')$ is an augmented triplet, then $y_{e'}$ is set to its weight (see Section~\ref{sec:applying-rules},$(2)$).
If the triplet is neither in the KG nor an augmented triplet, then $y_{e'} = 0$.


\subsection{Results}
\label{sec:results}

The results of the KGC experiments are shown in Table~\ref{kgcresult}.
We can see Transductive Augmentation achieves higher MRR and Hits@10 than Base and Random Augmentation.

The poor results of Random Augmentation indicate that it does not generate useful data for answering test queries,
implying the importance of selecting triplets to add to the training sets.
Transductive Augmentation appears to work well in terms of this selection, in particular, on WN18RR,
it even outperforms UniKER,
although comparison with UniKER is not fair as the latter is an inductive method and does not assume a transductive setting.
On FB15k-237, the performance of Transductive Augmentation is not so distinct compared with two other approaches and lower than UniKER.


\section{Analysis}
\label{sec:analysis}

\begin{table}[t]
  \caption{\label{validqueries}The numbers of the three types of validation queries}
  \centering
  \small
  \begin{tabular}{l r r r} \toprule
    Dataset   & all    & w/aug   & w/true \\ \midrule
    WN18RR    & 6,068  & 3,924   & 2,499  \\ 
    FB15k-237 & 35,070 & 13,005  & 6,163  \\ \bottomrule
  \end{tabular}
\end{table}

\begin{table}[t]
  \caption{\label{minedpath}The average and SD of the number of mined paths for each relation }
  \centering
  \small
  \begin{tabular}{l r r} \toprule
    Dataset                     & WN18RR      & FB15k-237   \\ \midrule
    \# mined paths per relation & 28k$\pm$26k & 29k$\pm$14k \\ \bottomrule
  \end{tabular}
\end{table}

In this section, we analyze the performance of the proposed method in more detail.
Throughout the section, TuckER is used as the embedding model.


\subsection{Poor performance on FB15k-237}

We first analyze the possible reason that Transductive Augmentation fails to perform as well on FB15k-237 as on WN18RR.
Table~\ref{validqueries} shows the numbers of three types of validation queries on two datasets:
``all'' represents all validation queries,
``w/aug'' represents queries that have augmented triplets, and
``w/true'' represents queries whose augmented triplets contain their true answers (i.e., the original triplets in the validation set before hiding one of its entities to make queries).
Thus, ``w/true'' is a subset of ``w/aug'', which is a subset of ``all''; i.e., $\text{``w/true''} \subseteq \text{``w/aug''} \subseteq \text{``all''}$.
Note that, for ``w/aug'' and ``w/true,'' the threshold values for data augmentation are those calibrated on the performance of the transductive augmentation with the TuckER embedding model.

According to Table~\ref{validqueries}, the ratio of the ``w/aug'' or ``w/true'' queries to ``all'' the queries on FB15k-237 is much smaller than on WN18RR.
This means that 
a smaller number of candidate entities (``w/aug'') and true answers (``w/true'') were generated for each query on FB15k-237.
This could be one of the reasons that lead to the worse results on FB15k-237.

See also Table~\ref{minedpath} for the average and standard deviation of the number of mined relation path rules.
The two datasets have nearly the same number of rules per relation.
FB15k-237 seems to have a smaller number of relation path rules, considering the number of relation types on FB15k-237 is twenty times larger than WN18RR.  

These analyses suggest that FB15k-237 has a few relation path rules and we have not been able to augment many data. 
If we expand the mining spaces of relation path rules at the expense of computational cost, the result of the proposed method on FB15k-237 might be improved.

\subsection{Improvements in the prediction of individual queries}


We now analyze how the prediction for individual queries differ between the models trained with the Base dataset (i.e., without data augmentation) and with the dataset augmented by Transductive Augmentation.
We call the two trained models simply as the Base and Transductive Augmentation models, respectively.
Tables~\ref{wnqueryh10} and \ref{fbqueryh10} show the breakdown of validation queries by whether or not the gold entities are ranked in the top 10 by each model on WN18RR and FB15k-237.

\label{sec:ImpOfPred}
\begin{table}[!t]
  \caption{\label{wnqueryh10}Breakdown of queries by the rankings of gold entities (WN18RR)}
  \centering
  \footnotesize
  \tabcolsep 3pt
  \begin{tabular}{r r r r @{\quad} r r r}
    \toprule
                        & \multicolumn{6}{c}{Base ranks}                                                                                                                                                            \\ \cmidrule{2-7}
                        & \multicolumn{3}{c}{gold in top-10} & \multicolumn{3}{c}{gold outside top-10}                                                                                                              \\ \cmidrule(r){2-4} \cmidrule(l){5-7} 
Trans Aug ranks         & all                                & \multicolumn{1}{l}{w/aug}    & \multicolumn{1}{l}{w/true}   & all                         & \multicolumn{1}{l}{w/aug}   & \multicolumn{1}{l}{w/true} \\ \midrule
    gold in top-10      & \cellcolor{RemainHigh} 2,979       & \cellcolor{RemainHigh} 2,527 & \cellcolor{RemainHigh} 2,404 & \cellcolor{Improved} 461    & \cellcolor{Improved} 206    & \cellcolor{Improved} 94    \\ \addlinespace[2pt]
    gold outside top-10 & \cellcolor{Degraded} 89            & \cellcolor{Degraded} 49      & \cellcolor{Degraded} 0       & \cellcolor{RemainLow} 2,539 & \cellcolor{RemainLow} 1,142 & \cellcolor{RemainLow} 1    \\ \bottomrule
  \end{tabular}
\end{table}

\begin{table}[!t]
  \caption{\label{fbqueryh10}Breakdown of queries by the rankings of gold entities (FB15k-237)}
  \footnotesize
  \centering
  \tabcolsep 3pt
  \begin{tabular}{r r r r @{\quad} r r r} 
    \toprule
                                        & \multicolumn{6}{c}{Base ranks}                                                                                                                                                             \\ \cmidrule{2-7}
                                        & \multicolumn{3}{c}{gold in top-10} & \multicolumn{3}{c}{gold outside top-10}                                                                                                               \\ \cmidrule(r){2-4}  \cmidrule(l){5-7}
                        Trans Aug ranks & all                                & \multicolumn{1}{l}{w/aug}    & \multicolumn{1}{l}{w/true}   & all                          & \multicolumn{1}{l}{w/aug}   & \multicolumn{1}{l}{w/true} \\ \midrule
    gold in top-10                      & \cellcolor{RemainHigh} 17,796      & \cellcolor{RemainHigh} 9,094 & \cellcolor{RemainHigh} 5,893 & \cellcolor{Improved} 1,421   & \cellcolor{Improved} 446    & \cellcolor{Improved} 136   \\ \addlinespace[2pt] 
    gold outside top-10                 & \cellcolor{Degraded} 1,189         & \cellcolor{Degraded} 338     & \cellcolor{Degraded} 24      & \cellcolor{RemainLow} 14,664 & \cellcolor{RemainLow} 3,127 & \cellcolor{RemainLow} 110  \\ \bottomrule
  \end{tabular}
\end{table}

In these tables, each query falls into one of the four types according to the ranks of gold entities by the Base and Transductive Augmentation (written `Trans Aug' in the tables) models: 
\begin{itemize}
\item \textit{Improved}:   queries for which the Base model ranked the gold entities outside the top-10 and Transductive Augmentation ranked them in the top-10 (the upper-right cells with blue background)
\item \textit{Degraded}:   queries for which the Base model ranked gold entities in the top-10 but Transductive Augmentation failed to do so (lower-left cells in red)
\item \textit{RemainLow}:  queries for which both Base and Transductive Augmentation ranked gold entities outside the top-10 (lower-right cells in yellow)
\item \textit{RemainHigh}: queries for which both Base and Transductive Augmentation ranked gold entities in the top-10 (upper-left cells in green)
\end{itemize}
In addition to the number of queries in individual categories shown under column ``all,'' the tables further show for each category the numbers of ``w/aug'' queries (i.e., those with at least one augmented triple), and
``w/true'' queries (i.e., those with gold triples in the augmented data), as defined in Section~\ref{sec:analysis}.

From Tables~\ref{wnqueryh10} and \ref{fbqueryh10}, we see that both datasets WN18RR and FB15k-237 exhibit the following similar trends.

First, by comparing `all' columns, we see that, overall, there are more \textit{Improved} queries (upper-right cells in blue) than \textit{Degraded} queries (lower-left cells in red).
This has led Transductive Augmentation outperforming the Base model, as shown in Section~\ref{sec:results}.
This trend also applies to `w/aug' and `w/true' subcategories, for which we see more \textit{Improved} than \textit{Degraded} queries.
Thus, having augmented data generally improves prediction accuracy.

However, the tables 
also show that a large number of the w/aug and w/true queries fall into the \textit{RemainHigh} category.
For these augmented triplets, the information they provide is already captured by the Base embedding models. 

Some w/aug and w/true queries fall into the \textit{RemainLow} category.
This category shows that Transductive Augmentation fails to add informative triples to the training set, either because there are no applicable relation path rules or their confidence weight is low.


\begin{table}[t]
  \caption{\label{cossim}Cosine similarity MRR of three types of queries}
  \centering
  \footnotesize
  \tabcolsep=.6em
  \begin{tabular}[]{l r r r r} \toprule
                       & \multicolumn{2}{c}{WN18RR} & \multicolumn{2}{c}{FB15k-237}            \\
    \cmidrule{2-3} \cmidrule(l){4-5}
                       & Similarity MRR             & \#triplets & Similarity MRR & \#triplets \\ \midrule
    \textit{Improved}  & \textbf{0.146}             & 114        & \textbf{0.102} & 310        \\
    \textit{Degraded}  & 0.0214                     & 47         & 0.0677         & 314        \\
    \textit{RemainLow} & 0.0237                     & 1137       & 0.0447         & 3017       \\ \bottomrule
  \end{tabular}
\end{table}


\subsection{Cosine similarity of augmented data}

We now analyze queries for which Transductive Augmentation added matching triplets but failed to add correct answers (gold triplets).
Although gold triplets were missing in the augmented data, the prediction on some of these queries improved.
We analyze this improvement by the models' performance on the validation set.
Specifically, we verify whether it owes to augmented triplets that are not exactly the gold triplets but contain similar entities to the one in the gold.

Let $(e_h,r,?)$ be one such query, i.e., it has matching augmented triplets but none of them is the gold triplet $(e_h, r, e_t)$ from which the query was generated.
In other words, $e_t$ is the correct answer for this query.
Further let $E_t(e_h,r)$ be the set of tail entities in the triplets augmented for the query.

For this query, we generate a ranking list by sorting all entities $E$ in the KG by their similarity to the gold entity $e_t$, using the cosine of two embeddings as the similarity measure.
If there are other gold entities for query $(e_h, r, ?)$ in the dataset, we remove them from the ranking list except for those appearing in $E_t(e_h,r)$;
this is similar to the standard ``filtered'' setting used in Section~\ref{sec:exp}.
Using the resulting ranking list, we define the ``similarity rank'' of $E_t(e_h,r)$ with respect to $e_t$ as the highest rank among the entities $E_t(e_h,r)$.

We compute the similarity rank for each validation query for which augmented triplets exist but do not include gold triplets.
For analysis, we divide them into three types defined in Section~\ref{sec:ImpOfPred}, \textit{Improved}, \textit{Degraded}, and \textit{RemainLow},
and compute their MRR (``similarity MRR'') individually.

The resulting ``similarity MRRs'' are shown in Table~\ref{cossim} for each type along with the numbers of triplets.
Queries in the \textit{Improved} category has the highest similarity MRR on both the WN18RR and FB15k-237 datasets, suggesting that learning similar entities to those in true answers contributed to improving the prediction.


\section{Conclusions}

In this paper, we have proposed a KGE approach that uses mined relation path rules for data augmentation, which is capable of dealing with relation path rules with low confidence scores.
This is contrasting to the previous embedding/rule hybrid model UniKER, which only utilizes high-confidence rules.
To augment data relevant to test queries, we assumed a transductive setting, and showed its effectiveness;
the KGC performance of the embedding models were improved by the transductive augmentation, and they outperformed UniKER's inductive approach on WN18RR.

In future work, we plan to apply our method to other KGE models than those used in this paper.
We can also incorporate other data augmentation approaches such as noise reduction used in UniKER.
Transductive information could be useful for mining relation path rules only to predict test triplets because the mining search space of KG is very large.
No previous studies considered transductive settings for KGC, which should have a large room for further investigation. 


\section*{Acknowledgments}
\addcontentsline{toc}{section}{Acknowledgment}

MS was partially supported by JSPS Kakenhi Grant 19H04173.


\appendices

\begin{figure}[t]
  \vspace{0cm}
  \begin{algorithm}[H]
    \footnotesize
    \caption{Mining path of the rules to predict relation $r$}
    \label{alg1}
    \begin{algorithmic}[1]
      \baselineskip=1.2\baselineskip
      \STATE Randomly sample $\mathit{sampleNum}$ triplets for $r$ and put them in $\mathit{Triplets}$
      \STATE $\mathit{minedPath}=\phi$
      \FOR {\textbf{each} $(e_h, r, e_t) \in \mathit{Triplets}$}
      \STATE $SG_h = \phi$; $SG_t = \phi$
      \FOR{$l=1$ \textbf{to} $\mathit{maxLength}$}
      \FOR{$j=1$ \textbf{to} $\mathit{tryNum}$}
      \STATE $SG_h.add(\mathit{randomWalk}(e_h,l))$
      \STATE $SG_t.add(\mathit{randomWalk}(e_t,l))$
      \ENDFOR
      \ENDFOR
      \STATE $\mathit{minedPath}.\mathit{add}(\mathit{merge}(SG_h,SG_t))$
      \ENDFOR
      \RETURN $\mathit{minedPath}$

    \end{algorithmic}
  \end{algorithm}
\end{figure}


\section{Details of relation path rule mining}
\label{sec:rule-mining-details}

We mine relation paths rules to predict relation $r$ from training data
as shown in the pseudocode of Algorithm~\ref{alg1}.
The mining process can be summarized as follows.
\begin{enumerate}
\item Randomly sample $\mathit{sampleNum}$ triplets for relation $r$ from the input KG.
\item For each sampled triplet $(e_h,r,e_t)$, perform $\mathit{tryNum}$ random walks of length $\mathit{maxLength}$ starting from each of $e_h$ and $e_t$ on the KG.
  In Algorithm~\ref{alg1}, $\mathit{randomWalk}(e,l)$ is a function that outputs a path traversed by an $l$-step random walk from node $e$,
  i.e., this path has a form $(e_0=e, r_1, e_1, \ldots, r_l, e_l)$ where $\{e_i\}$ are entities and $\{r_i\}$ are relations.
  The traversed paths are stored in sets $SG_h$ and $SG_t$.
  After repeating this for $\mathit{tryNum}$ times, function $\mathit{merge}$ is called, which examines paths in $SG_h$ and $SG_t$ and if paths from the two sets have the same end nodes, they are concatenated.
  Function $\mathit{merge}$ further removes entities in the concatenated paths to produce relation paths,
  which are then accumulated in $\mathit{minedPath}$.

\end{enumerate}

We set $(\mathit{sampleNum}, \mathit{maxLength}, \mathit{tryNum})\allowbreak=\allowbreak(6000,\allowbreak 3,\allowbreak 10000)$
for WN18RR and $(100,\allowbreak 3,\allowbreak 300)$ for FB15k-237.
As a result, relation paths of length 1 to 6 are mined on both datasets.


\section{Hyperparameters for trained embedding models}
\label{sec:hyperparams}

Tables~\ref{tab:hyparawn} and \ref{tab:hyparafb} show
the hyperparameters of the embedding models used in the experiments.
These values were determined by the performance on the validation set, as described in Section~\ref{sec:exp-settings}.

\begin{table}[t]
  \caption{Hyperparameters for WN18RR}
  \label{tab:hyparawn}
  \hspace{0cm}
  \small
  \begin{tabular}[]{l c c} \toprule
    Dataset                                             & \multicolumn{2}{c}{WN18RR}                      \\ \midrule
    Embedding model                                     & RESCAL   & TuckER                               \\ \midrule
    Optimizer                                           & \multicolumn{2}{c}{Adam~\cite{adam}}            \\
    Learning rate                                       & 0.001    & 0.003                                \\
    Learning decay rate                                 & 1.       & 0.99                                 \\
    Input, hidden1, hidden2 dropout rates               & \multicolumn{2}{c}{(0.2,0.2,0.3)}               \\
    Batch size                                          & \multicolumn{2}{c}{128}                         \\
    Entity embedding dimension                          & \multicolumn{2}{c}{200}                         \\
    Relation embedding dimension                        & $200^2$  & 30                                   \\
    Label smoothing                                     & \multicolumn{2}{c}{0.1}                         \\
    Embedding initialization                            & \multicolumn{2}{c}{Xavier normal~\cite{xavier}} \\
    Max iteration                                       & \multicolumn{2}{c}{500}   
                                                                                                          \\
    Augmentation thresholds ($\topN$, $\confThold$) & 
                                                                                                          \\
    \qquad Random augmentation                          & (50,0.6) & (50,0)
                                                                                                          \\
    \qquad Transductive augmentation                    & (5,0)    & (5,0)
                                                                                                          \\ \bottomrule
  \end{tabular}
\end{table}

\begin{table}[t]
  \caption{Hyperparameters for FB15k-237}
  \label{tab:hyparafb}
  \hspace{0cm}
  \small
  \begin{tabular}[]{l c c} \toprule
    Dataset                                             & \multicolumn{2}{c}{FB15k-237}     \\ \midrule
    Embedding model                                     & RESCAL  & TuckER                  \\ \midrule
    Optimizer                                           & \multicolumn{2}{c}{Adam}          \\ 
    Learning rate                                       & 0.003   & 0.001                   \\
    Learning decay rate                                 & 0.995   & 1.                      \\
    Input, hidden1, hidden2 dropout rates               & \multicolumn{2}{c}{(0.3,0.4,0.5)} \\
    Batch size                                          & \multicolumn{2}{c}{128}           \\
    Entity embedding dimension                          & \multicolumn{2}{c}{200}           \\
    Relation embedding dimension                        & $200^2$ & 200                     \\
    Label smoothing                                     & \multicolumn{2}{c}{0.1}           \\
    Embedding initialization                            & \multicolumn{2}{c}{Xavier normal} \\
    Max iteration                                       & \multicolumn{2}{c}{500}           \\
    Augmentation thresholds ($\topN$, $\confThold$)     &                                   \\
    \qquad Random augmentation                          & (5,0)   & (5,0.6)                 \\
    \qquad Transductive augmentation                    & (5,0)   & (5,0)                 \\ \bottomrule
  \end{tabular}
\end{table}


\bibliographystyle{IEEEtran}
\bibliography{ref.bib}


\end{document}